\documentclass{article} 
\usepackage{iclr2023_conference,times}

\usepackage{amsmath,amsfonts,bm}

\def\eqref#1{equation~\ref{#1}}

\def\1{\bm{1}}

\DeclareMathAlphabet{\mathsfit}{\encodingdefault}{\sfdefault}{m}{sl}
\SetMathAlphabet{\mathsfit}{bold}{\encodingdefault}{\sfdefault}{bx}{n}

\usepackage{hyperref}
\usepackage{url}
\usepackage{booktabs}       
\usepackage{amsfonts}       
\usepackage{nicefrac}       
\usepackage{microtype}      
\usepackage{xcolor}         

\usepackage{amsmath}
\usepackage{multirow}
\usepackage{makecell}
\usepackage{colortbl}
\usepackage{graphicx}
\usepackage{subcaption}
\usepackage{alltt}
\usepackage{color}
\usepackage{algorithmicx,algorithm}
\usepackage{wrapfig}
\definecolor{Gray}{gray}{0.9}
\definecolor{red}{RGB}{215,48,31}
\definecolor{green}{RGB}{65,171,93}

\newcommand{\eg}{\textit{e.g.}}

\title{ViTKD: Practical Guidelines for ViT feature knowledge distillation}

\author{Zhendong Yang$^{1,2}$ \quad Zhe Li\quad Ailing Zeng$^{2}$\quad Zexian Li$^{3}$\quad Chun Yuan\textsuperscript{$\dagger$}$^{1}$\quad Yu Li\textsuperscript{$\dagger$}$^{2}$\\
$^{1}$Tsinghua Shenzhen International Graduate School\\
$^{2}$International Digital Economy Academy  (IDEA) \quad$^{3}$Beihang University\\
{\tt\small \{yangzd21@mails, yuanc@sz\}.tsinghua.edu.cn \quad axel.li@outlook.com}\\
{\tt\small \{zengailing, liyu\}@idea.edu.cn\quad lizexian0427@gmail.com}
}

\iclrfinalcopy 
\begin{document}

\maketitle
\renewcommand{\thefootnote}{\fnsymbol{footnote}} 
\footnotetext[2]{Corresponding authors} 

\begin{abstract}

Knowledge Distillation (KD) for Convolutional Neural Network (CNN) is extensively studied as a way to boost the performance of a small model. Recently, Vision Transformer (ViT) has achieved great success on many computer vision tasks and KD for ViT is also desired. However, besides the output logit-based KD, other feature-based KD methods for CNNs cannot be directly applied to ViT due to the huge structure gap. In this paper, we explore the way of feature-based distillation for ViT. Based on the nature of feature maps in ViT, we design a series of controlled experiments and derive three practical guidelines for ViT's feature distillation. Some of our findings are even opposite to the practices in the CNN era. Based on the three guidelines, we propose our feature-based method ViTKD which brings consistent and considerable improvement to the student. On ImageNet-1k, we boost DeiT-Tiny from $74.42\%$ to $76.06\%$, DeiT-Small from $80.55\%$ to $81.95\%$, and DeiT-Base from $81.76\%$ to $83.46\%$. Moreover, ViTKD and the logit-based KD method are complementary and can be applied together directly. This combination can further improve the performance of the student. Specifically, the student DeiT-Tiny, Small, and Base achieve $77.78\%$, $83.59\%$, and $85.41\%$, respectively.
\footnote{The code is available at \url{https://github.com/yzd-v/cls_KD}.}  

\end{abstract}

\section{Introduction}
\label{intro}
Knowledge Distillation (KD) ~\citep{hinton2015distilling} utilizes the output of the teacher model as soft labels to supervise the student model, bringing the lightweight models impressive improvements without extra costs for inference. It has been consistently explored for Convolutional Neural Network (CNN) models and applied successfully to many vision tasks successfully, including image classification~\citep{zhou2020rethinking,yang2020knowledge,chen2021distilling,zhao2022decoupled,lin2022knowledge}, object detection~\citep{li2022knowledge,yang2022focal,zheng2022localization,yang2022prediction,wang2022head}, semantic segmentation~\citep{liu2019structured,he2019knowledge,shu2021channel,yang2022cross}.

Recently, Vision Transformer (ViT) ~\citep{dosovitskiy2021an} has achieved great success for image classification and inspired various transformers~\citep{yuan2021tokens,han2021transformer,Touvron_2021_ICCV,liu2021swin}. Compared with CNN-based models, the ViT-based methods generally need more parameters but can achieve better performance, making them harder to be deployed. Therefore, boosting the performance of small ViT models using KD is of great value. 
In this work, we look into \emph{how to apply KD to ViT-based models.} A direct thought would be directly transferring the KD methods for CNN to ViT. In fact, some fundamental distillation works~\citep{hinton2015distilling,romero2014fitnets} are inherently structure-independent. For example, the classic logit-based distillation directly use the model's final output logit, thus it can apply for both CNNs and ViTs. DeiT~\citep{touvron2021training} verifies this for ViT's distillation.

However, the rest KD methods are mostly specially designed for CNN-based models and many of them work on the intermediate features. They are inapplicable to ViT-based models as there is a huge gap between these two architectures. The recent MiniViT~\citep{zhang2022minivit} adopts Self-Attention distillation and Hidden-State Distillation for feature-based distillation. Compared with the logit based distillation, its improvement is still quite limited.

Before developing new feature-based KD for ViT, we first conduct simple studies of transferring the knowledge from the last layer of a teacher (DeiT-Small) following the CNN's distillation, from the last 6 layers like PKD~\citep{sun2019patient} for BERT's~\citep{devlin2018bert} distillation, and from the whole 12 layers. Surprisingly, the results for all the intuitive feature distillations shown in Table~\ref{table:simple fea dis} are not satisfactory which consistently degrade the performance of the student (DeiT-Tiny). Specifically, the Top-1 accuracy of the student is just $73.36\%$ when distilling on the last layer. This distillation on the last layer is widely used for CNN's distillation, but here it causes a $1.06\%$ accuracy drop. 

\begin{table}
\setlength{\tabcolsep}{12.5 pt}
  \centering
  \caption{The results of different feature distillation methods, including distillation on the last layer like CNN's distillation, the last 6 layers like BERT's distillation, and all the 12 layers.}
  \begin{tabular}{c|c|ccc}
    \toprule
    Distillation setting & \multicolumn{4}{c}{DeiT Small (Teacher) - DeiT Tiny (Student) }\\
    \midrule
    Methods&Baseline& Last layer&Last 6 layers &All 12 layers\\
    \midrule
    Top-1 Accuracy & 74.42 &73.36 ({\color{green}-1.06})&73.76 ({\color{green}-0.66})&74.24 ({\color{green}-0.18})\\
    Top-5 Accuracy & 92.29 &91.88&92.01&92.23\\
    \bottomrule
  \end{tabular}
  \label{table:simple fea dis}
\end{table}

\begin{figure}[t]
    \centering
    \includegraphics[width=\textwidth]{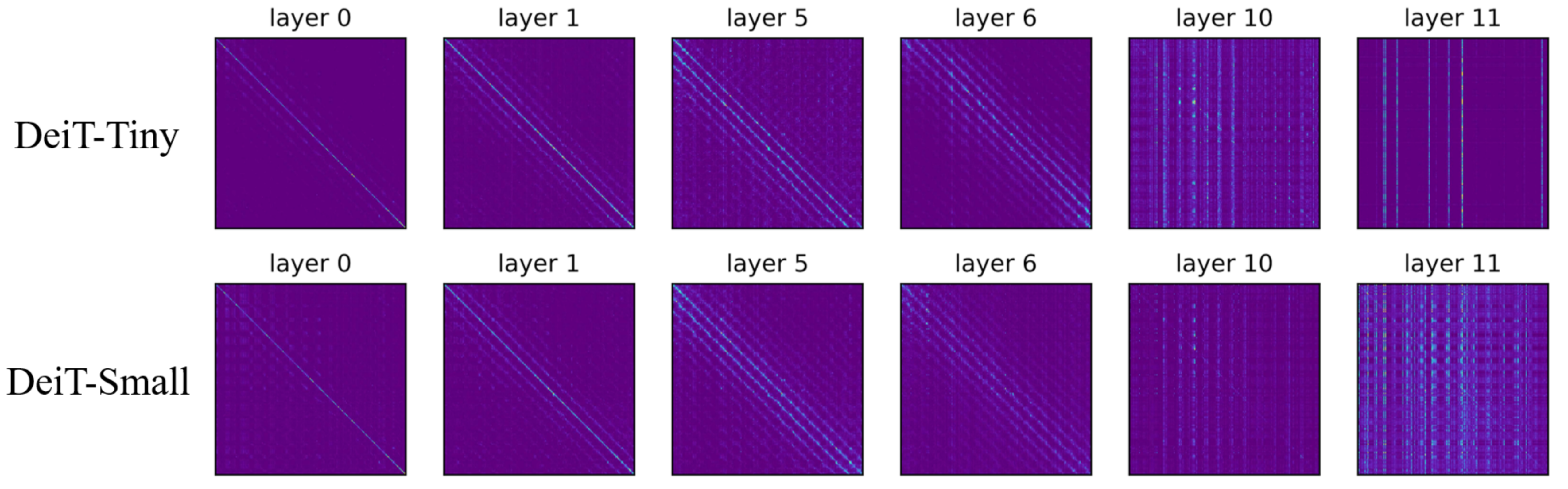}
    \caption{DeiT-Tiny's and DeiT-Small's attention maps from shallow to deep layers. The X-axis and Y-axis mean the key and query tokens, respectively. The attention map is obtained by \emph{softmax} and reflects the response between the query and key tokens. The color is brighter with a larger response.}
    \label{fig:attn_tiny}
\end{figure}

\setlength{\tabcolsep}{2pt}
\begin{wrapfigure}{r}{0.5\textwidth}
    \centering
    \includegraphics[width=0.45\textwidth]{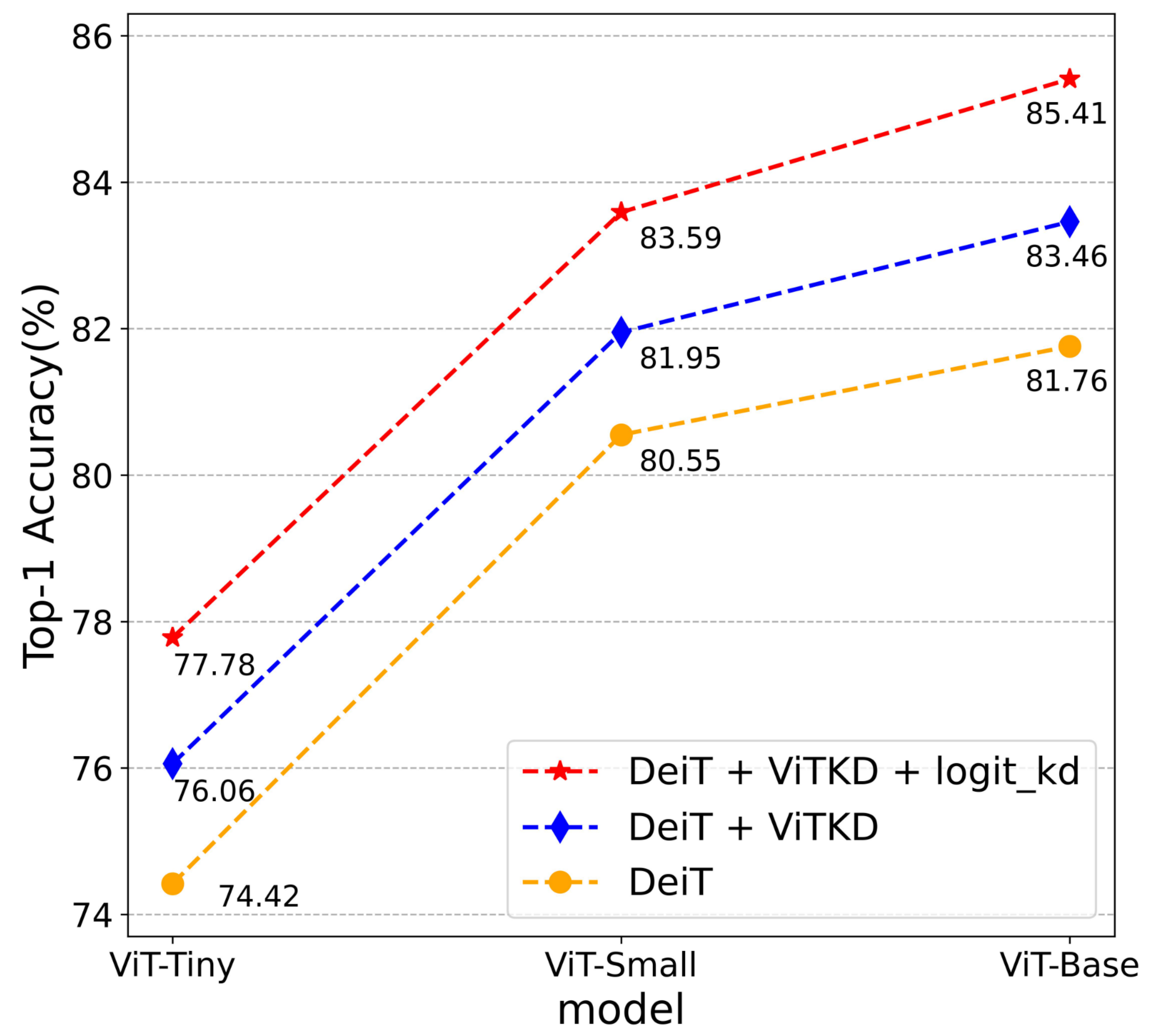}
    \caption{Comparison of training vision transformers with distillation on ImageNet-1k.}
    \label{fig:line}
\end{wrapfigure}

To further explore the features in ViT, we visualize the attention maps of the student and the teacher across different layers in Figure~\ref{fig:attn_tiny}. For the shallow layers (\eg, layers 0 and 1), the attention appears mainly on the diagonal, which indicates they focus on themselves. Both the student and teacher have similar patterns. While for the deep layers (\eg, layers 10 and 11), the difference between student and teacher's attention is greater. Their attention is decided by several sparse key tokens. Besides, they focus on completely different tokens. Such gap makes it hard for the student to mimic the teacher's final feature directly. This phenomenon suggests different layers may need different methods.

Accordingly, we perform a series of controlled experiments to examine the effects of different distillation methods, different layers, and different modules. As a consequence, we derive three practical guidelines for ViT's feature distillation in Section~\ref{guidelines}. Based on these principles, we propose a nontrivial way for feature-based ViT distillation, named~\textbf{ViTKD}, and describe the details in Section~\ref{sec:method}. Extensive experiments demonstrate its effectiveness in Section~\ref{experiment}. For instance, we boost the student DeiT-Tiny from $74.42\%$ to $76.06\%$, DeiT-Small from $80.55\%$ to $81.95\%$ and DeiT-Base from $81.76\%$ to $83.46\%$ on ImageNet-1K. Besides, when combining ViTKD with the logit-based distillation, we can further advance their Top-1 accuracy to $77.78\%$, $83.59\%$ and $85.41\%$. The comparison is shown in Figure~\ref{fig:line}.  We also demonstrate the models trained with distillation are beneficial to other vision tasks like object detection.

\section{Practical Guidelines for ViT's Feature Distillation}
\label{guidelines}

To explore the practical guidelines, we take larger DeiT~\citep{touvron2021training} and DeiT III~\citep{touvron2022deit} models as the teacher to distill lighter DeiT models on ImageNet-1k~\citep{deng2009imagenet}. The DeiT teacher is trained from scratch on ImageNet-1K, and DeiT III teacher is pre-trained on ImageNet-21K. As shown in Section~\ref{intro}, the attention maps vary greatly from different layers. Based on this observation, we analyze where and how to distill the student effectively and propose three practical guidelines for ViT's feature distillation. Specifically, we conduct distillation experiments on features in different layers of DeiT with two strategies, namely \emph{mimicking} and \emph{generation}. When using \emph{mimicking}, we align the embedding dimensions of the student and the teacher by a linear layer and correlation matrix, respectively. As for \emph{generation}, we randomly mask the student's tokens and utilize a generative block to restore the feature. Furthermore, we choose different generative blocks, including cross-attention block~\citep{chen2022context}, self-attention block~\citep{he2022masked}, and convolutional projector~\citep{yang2022masked}. For both \emph{mimicking} and \emph{generation}, we calculate the square of $L_2$ distance as the distillation loss. The details about \emph{mimicking} and \emph{generation} are elaborated in Subsection~\ref{sec:mimicking} and~\ref{sec:generation}, respectively.

\begin{table*}[t]
\setlength{\tabcolsep}{14 pt}
  \centering
  \caption{The comparisons of Top-1 accuracy with different distillation methods on DeiT's deep-layer feature on ImageNet-1K.}
  \begin{tabular}{c|c|c|c}
    \toprule
    \multirow{2}{*}{\makecell{Type}}&Teacher & DeiT-Small (80.69) &DeiT III-Small (82.76)\\
    &Student & DeiT-Tiny (74.42) &DeiT-Tiny (74.42)\\
    \midrule
    \multirow{2}{*}{\makecell{\emph{Mimicking}}}
    &Linear layer & 73.36 ({\color{green}-1.06})&73.72 ({\color{green}-0.70})\\
    &Correlation matrix & 72.37 ({\color{green}-2.05})&72.20 ({\color{green}-2.22})\\
    \midrule
    \multirow{3}{*}{\makecell{\emph{Generation}}}
    & Cross-attention &73.77 ({\color{green}-0.65})& 73.98 ({\color{green}-0.44})\\
    & Self-attention &74.61 ({\color{red}+0.19})& 74.65 ({\color{red}+0.23})\\
    & Conv. projector &\textbf{74.72} ({\color{red}+0.30})&\textbf{74.79} ({\color{red}+0.37})\\
    \bottomrule
  \end{tabular}
  \vspace{2pt} 
  \label{table:g1}
\end{table*}

{\bf G1) For distillation on the deep layer, \emph{generation} is more suitable than \emph{mimicking} .} For CNN's feature distillation, many works~\citep{park2019relational,tian2019contrastive,yang2022masked} transfer teachers' semantic information from the last-stage feature. Most feature-based distillation methods~\citep{romero2014fitnets,zagoruyko2016paying,heo2019comprehensive} aim at making students get similar feature maps to the teacher. While MGD~\citep{yang2022masked} forces the student to generate the teacher's full feature instead of mimicking it directly.

As our results shown in Table~\ref{table:g1}, the way to mimic the last layer feature of the teacher surprisingly impair the student's performance noticeably. Specifically, the student's Top-1 accuracy drops about $2\%$ when mimicking the teacher's correlation matrix. This trend is completely different from distillation for CNN-based models. Instead, the generation methods can improve the accuracy of the student mostly. The largest gains are obtained by using  the convolutional projector as the generative block. These results reveal that \emph{generation} is more suitable than \emph{mimicking} for the deep layer.

{\bf G2) Distillation on the shallow layers also works for ViT with \emph{mimicking}.} For the CNN-based model's feature distillation, the feature of shallow layers has a small receptive field and lacks semantic information, making it unsuitable for distillation. As the attention map shown in Figure~\ref{fig:attn_tiny}, the shallow feature of DeiT also has a small receptive. That is, the tokens in the first two layers just have responses to themselves. Still, we believe such incipient attention knowledge is useful for distillation because it can teach the student how to form a better attention map at the beginning.

We pick the first two layers for distillation by either mimicking or generation in Table~\ref{table:g2}. Interestingly, the conclusion for feature distillation on the shallow layers and deep layers is the opposite. The relations of different tokens and semantic information from the shallow layers are so weak that the student can not utilize its masked feature to generate the full feature from the teacher. It makes the generation way just bring a little improvement for the shallow-layer distillation. 
Moreover, different from distillation for CNN-based models, transferring the knowledge from teacher's shallow layer by directly mimicking makes great progress. 
Mimicking by `correlation matrix' performs a little better than the `linear layer' way when using DeiT-S as the teacher. When the teacher performs better, the `linear layer' way benefits the student much more than the `correlation matrix' way. 
As we described above, the results validate that the shallow layer matters for distillation by mimicking. We fix to use the `linear layer' strategy to distill the shallow layers.

\begin{table*}[t]
\setlength{\tabcolsep}{9 pt}
  \centering
  \caption{The comparisons of Top-1 accuracy with different distillation methods on DeiT's shallow-layer feature on ImageNet-1k.}
  \begin{tabular}{c|c|ccc}
    \toprule
    \multirow{2}{*}{\makecell{Type}}  &Teacher & DeiT-S (80.69) &DeiT III-S (82.76) &DeiT-B (81.76)\\
    &Student & DeiT-T (74.42) &DeiT-T (74.42)&DeiT-T (74.42)\\
    \midrule
    \multirow{2}{*}{\makecell{\emph{Mimicking}}}
    &Linear layer & 75.12 ({\color{red}+0.70}) &\textbf{75.31} ({\color{red}+0.89})&\textbf{75.15} ({\color{red}+0.73})\\
    &Correlation matrix & \textbf{75.27} ({\color{red}+0.85}) &74.94 ({\color{red}+0.52}) &75.01 ({\color{red}+0.59})\\
    \midrule
    \emph{Generation}& Conv. projector &74.69 ({\color{red}+0.27}) &74.86 ({\color{red}+0.44})&74.71 ({\color{red}+0.29})\\
    \bottomrule
  \end{tabular}
  \vspace{2pt} 
  \label{table:g2}
\end{table*}

\begin{table*}[t]
\setlength{\tabcolsep}{12 pt}
  \centering
  \caption{The comparisons of Top-1 accuracy with different distillation modules on DeiT's shallow layer feature on ImageNet-1k.}
  \begin{tabular}{c|c|cc}
    \toprule
    \multirow{2}{*}{\makecell{Type}}&Teacher & DeiT-Small (80.69) &DeiT III-Small (82.76)\\
    &Student & DeiT-Tiny (74.42) &DeiT-Tiny (74.42)\\
    \midrule
    \multirow{2}{*}{\makecell{Modules}}
    &MHA & 75.06 ({\color{red}+0.64})& 75.02 ({\color{red}+0.60})\\
    &FFN & \textbf{75.12} ({\color{red}+0.70})& \textbf{75.31} ({\color{red}+0.89})\\
    \bottomrule
  \end{tabular}
  \vspace{2pt} 
  \label{table:g3}
\end{table*}

{\bf G3) The FFN-out features are better than the MHA-out features for distillation.}
The ViT-based models are built by stacking several encoder layers. Each encoder layer consists of a multi-head attention (MHA) module and a feed-forward network (FFN) module. Based on the findings from {\bf G1} and {\bf G2}, we further conduct experiments on the first two layers of the student to explore how to choose the modules for ViT's feature distillation.

We use the `linear layer' way for the shallow-layer distillation on the MHA-out feature and FFN-out feature, respectively. 
Table~\ref{table:g3} demonstrates that distilling on the MHA-out feature or FFN-out both bring the student improvements and transferring the knowledge from the FFN-out feature is better than the MHA-out feature.

\section{Methodology}
\label{sec:method}
As mentioned in our guidelines in Section~\ref{guidelines}, we apply  `linear layer' and `correlation matrix' for \emph{mimicking}. For \emph{generation}, we use `cross-attention', `self-attention', and `conv. projector'. In this section, we describe the details of these methods and the final formulation of our ViTKD. To begin, we recall the feature distillation of CNN is based on the $L_2$ distances between the feature maps. We also follow this paradigm. The general form of CNN's feature distillation loss is as following:
\begin{equation}
    \mathcal{L}_{fea}=\sum_{k=1}^{C}\sum_{i=1}^{H}\sum_{j=1}^{W}\big( \mathcal{F}_{k,i,j}^{T}-f(\mathcal{F}_{k,i,j}^{S})\big)^{2},
  \label{general_feature_loss}
\end{equation}
where $\mathcal{F}^{T}$ and $\mathcal{F}^{S}$ denote the feature from the teacher and student, respectively, and $f(\cdot)$ is the adaptation layer to reshape the $\mathcal{F}^{S}$ to the same dimension as $\mathcal{F}^{T}$. $H$ and $W$ denote the height and width of the feature, and $C$ is the channel length. In the next, we shift to feature distillation for ViT.

\subsection{Mimicking for shallow layers}
\label{sec:mimicking}
For each sample, we can denote student's and teacher's feature as $\mathcal{F}^{S} \in \mathcal{R}^{ N\times D_{S}}$ and $\mathcal{F}^{T} \in \mathcal{R}^{ N\times D_{T}}$, respectively. For the mimicking method, we utilize a linear layer to align the dimension of the student's $D_{S}$ and the teacher's $D_{T}$. We term the strategy as `linear layer' and summarize it as:
\begin{equation}
    \mathcal{L}_{lr}=\sum_{i=1}^{N}\sum_{j=1}^{D}\big( \mathcal{F}_{i,j}^{T}-fc(\mathcal{F}^{S})_{i,j}\big)^{2},
  \label{eq:linear loss}
\end{equation}
where $fc(\cdot)$ is a linear layer to reshape the $\mathcal{F}^{S}$ to the same dimension as $\mathcal{F}^{T}$. $N, D$ denote the number of patch tokens and the embedding dimension of the teacher's feature.

Besides, we use a correlation matrix to describe the response among different patch tokens and force the student to learn the correlation matrix of the teacher's features. In this case, we do not need the adaption layer to align the embedding dimension. The correlation matrix for each sample can be calculated as:
\begin{equation}
    \mathcal{M} = \frac{\mathcal{FF}^{Tr}}{\sqrt{D}},
\end{equation}
where $\mathcal{F} \in \mathcal{R}^{N\times D}$ denotes the student or teacher's feature. $D$ is their embedding dimension and $Tr$ denotes transposition for the feature, so $\mathcal{F}^{Tr} \in \mathcal{R}^{D\times N}$. In this case, student's and teacher's relation matrices have the same shape $\mathcal{M} \in \mathcal{R}^{N\times N}$ and describe the response between different patch tokens. With `correlation matrix', we calculate the distillation loss as:
\begin{equation}
    \mathcal{L}_{rm}=\sum_{i=1}^{N}\sum_{j=1}^{N}\big( \mathcal{M}_{i,j}^{T}-\mathcal{M}_{i,j}^{S}\big)^{2}.
  \label{eq:relation loss}
\end{equation}

\subsection{Generation for deep layers}
\label{sec:generation}
For \emph{generation}, we first use a linear layer to align the feature dimension of the student and teacher. Then, we set a random mask $Mask\in \mathcal{R}^{N\times 1}$ and use masked tokens to replace the student's original tokens, which can be formulated as:
\begin{equation}
    \label{eq:mask}
    \mathcal{\hat F}_{i}^{S}=
    \begin{cases}
        masked~~token, & \text{if}\ r_{i}< \lambda \\ 
        original~~token, & \text{Otherwise},
    \end{cases}
\end{equation}
\begin{equation} 
    Mask_{i}=
    \begin{cases}
        1, & \text{if}\ r_{i}< \lambda \\ 
        0, & \text{Otherwise},
    \end{cases}
\end{equation}
where $r_{i}$ is a random number uniformly distributed in $[0,1]$ and $i \in [0, N-1]$ is the coordinates of the tokens dimension. $\lambda$ is a hyper-parameter that is set as 0.5 for all the experiments. The $masked~~token$ is the parameter to learn during training.

Finally, we use the new masked feature $\mathcal{\hat F}_{i}^{S}$ to generate teacher’s the full feature through a generative block $\mathcal{G}$, which can be formulated as follows:
\begin{equation}
    \mathcal{G}(\mathcal{\hat F}^{S})\longrightarrow \mathcal{F}^{T}.
  \label{generation method}
\end{equation}
 
 We choose three ways to set the generative block $\mathcal{G}$. The first way is a `cross-attention' block from CAE~\citep{chen2022context}, which includes 6 transformer layers. The second way is a `self-attention' block from MAE~\citep{he2022masked}, which also includes 6 transformer layers. The difference between them is that cross-attention uses masked tokens as the query tokens. The third way is a `convolutional projector' from MGD~\citep{yang2022masked}, which includes two conventional layers. For all the three ways, we only calculate the distillation loss of masked tokens.
 
 For \emph{generation} method, we design the distillation loss $\mathcal{L}_{gen}$ as:
\begin{equation}
    \mathcal{L}_{gen}=\sum_{i=1}^{N}\sum_{j=1}^{D}Mask_{i}\big( \mathcal{F}_{i,j}^{T}-\mathcal{G}( \mathcal{\hat F}_{i,j}^{S})\big)^{2}.
  \label{eq:gen loss}
\end{equation}

\begin{figure}[t]
    \centering
    \includegraphics[width=\textwidth]{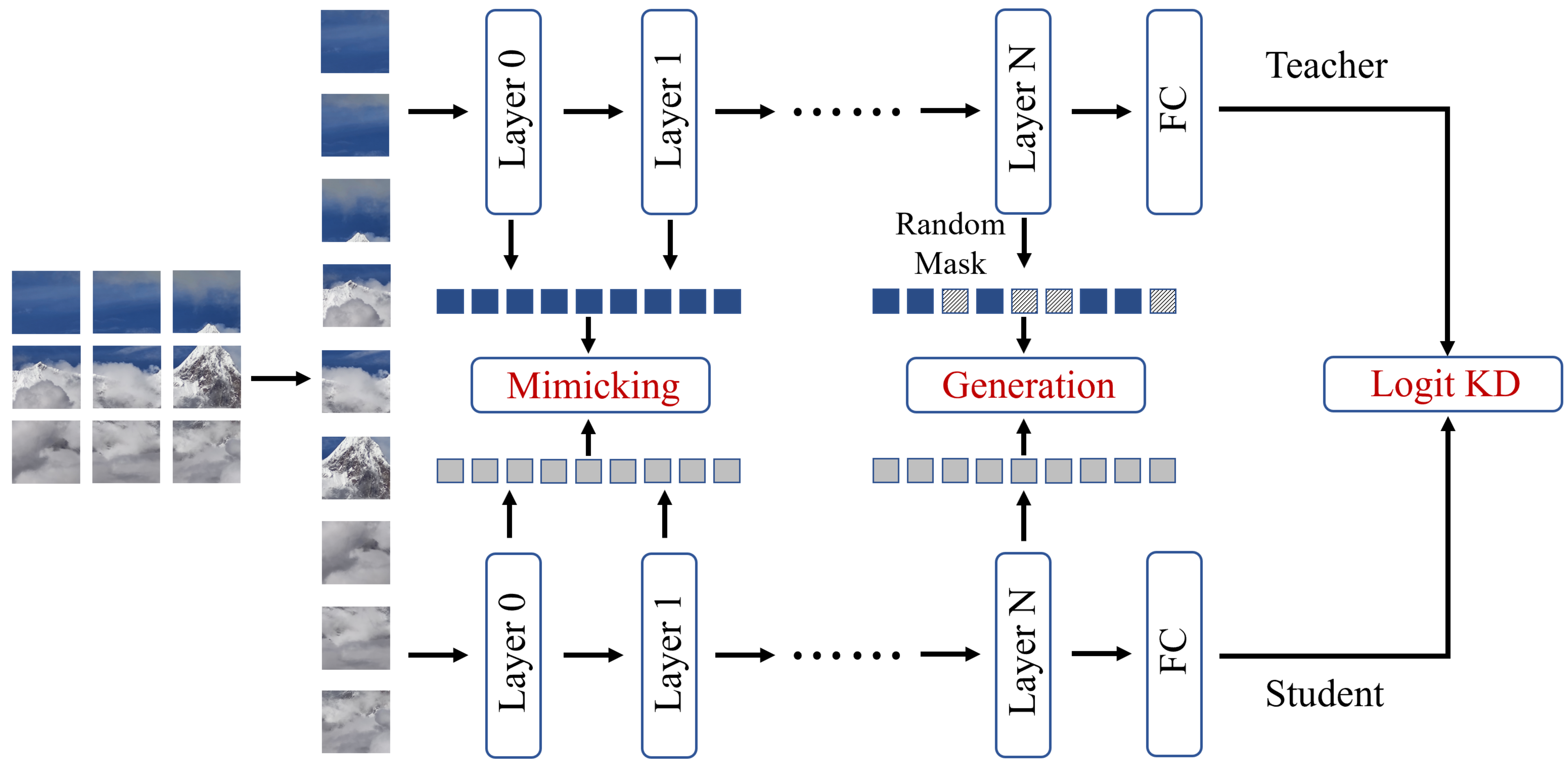}

    \caption{Illustration of the proposed ViTKD. ViTKD is a feature-based distillation method that includes \emph{Mimicking} and \emph{Generation}. It can be directly combined with the output logit-based distillation method together.}
    \label{fig:structure}
\end{figure}

\subsection{ViTKD}
\label{ViTKD}

Based on the findings from {\bf G1}, {\bf G2} and  {\bf G3}, we finally propose our method ViTKD. We first use the \textbf{`linear layer'} approach for the first two layers' distillation, where the distillation loss is $\mathcal{L}_{lr}$. As for the last layer, we apply the  \textbf{`conv. projector'} for the generation distillation, where the distillation loss is $\mathcal{L}_{gen}$. The ViTKD we propose is shown in Figure~\ref{fig:structure}. To sum up, we train the student model with the total loss as follows:
\begin{equation}
    \label{eq:all loss}
        \mathcal{L}=\mathcal{L}_{ori}+\alpha \mathcal{L}_{lr}+\beta \mathcal{L}_{gen},
\end{equation}
where $\mathcal{L}_{ori}$ is the original loss for the models, \eg, the cross-entropy loss in DeiT-Tiny. $\alpha$ and $\beta$ are two hyper-parameters to balance the loss.

\section{Experiment}
\label{experiment}

\subsection{Settings}
{\bf Datasets.} We explore the feature distillation for ViT-based models on ImageNet-1k~\citep{deng2009imagenet}, which contains 1000 object categories. We use the 1.2 million images to train the model and 50k images to evaluate the performance. For the downstream task, we evaluate our model on the COCO dataset~\citep{lin2014microsoft}, which contains 80 object classes. We use the 120k train images for training and 5k validation images for testing.

{\bf Implementation details.} ViTKD uses the hyper-parameters $\alpha$ and $\beta$ to balance the distillation loss in Equation~\ref{eq:all loss}. Another hyper-parameter $\lambda$ is used to adjust the masked ratio for deep layer distillation in Equation~\ref{eq:mask}. We adopt the hyper-parameters $\{\alpha = 3 \times 10^{-5}, \beta = 3 \times 10^{-6}, \lambda=0.5\}$ for all the experiments. As for the logit distillation, we apply the distillation method NKD~\citep{yang2022rethinking} and set the hyper-parameters $\{\alpha = 1, temperature=1\}$. Besides, to keep the model to be the same for the feature and logit distillation, we remove the extra distillation token which is used for logit distillation in DeiT. The image resolution for all the experiments is 224$\times$224. The other training details for distillation follow the setting from the baseline training setting in MMClassification~\citep{2020mmclassification}. All the experiments are conducted on 8 GPUs with MMClassification in Pytorch~\citep{paszke2019pytorch}. Unless specified, we evaluate the model with the performance of the last epoch.

\subsection{Main results}
To evaluate our methods for ViT-based models, we utilize different teachers to distill different students. We train the student with the proposed ViTKD, the classic KD, the state-of-the-art logit method NKD~\citep{yang2022rethinking}. We also combine ViTKD and NKD to explore the upper bound of the student's performance. In Table~\ref{table:main}, all the teachers bring the students remarkable performance improvements, \eg, the DeiT III-Small teacher boosts the student's Top-1 accuracy from $74.42\%$ to $76.06\%$ with our ViTKD method. The results of ViTKD even surpass the classic logit-based KD method. Comparing the results between different teachers, we find the student achieves better performance with a stronger teacher, \eg, the student DeiT-Tiny achieves $75.40\%$ and $76.06\%$ Top-1 accuracy with teh DeiT-Small and DeiT III-Small teacher, respectively. Furthermore, we also apply our method to a stronger student DeiT-Small and DeiT-Base. ViTKD can also bring them significant improvements, helping it to achieve $81.95\%$  and $83.46\%$, respectively.

Besides, ViTKD is a feature-based knowledge distillation method and can be combined with other logit-based methods for image classification. Therefore, we try to add the state-of-the-art logit-based distillation loss NKD to our ViTKD. In this way, the students with different teachers all get another significant accuracy improvement, \eg, the student DeiT-Small gets another $1.64\%$ gains and achieves $83.59\%$ Top-1 accuracy with a DeiT III-Base teacher. Surprisingly, the student DeiT-Small is just trained on ImageNet-1K, but its performance surpasses DeiT III-Small, which needs to be pre-trained on ImageNet-21k and then finetuned on ImageNet-1k.

\begin{table*}[t]
  \setlength{\tabcolsep}{10 pt}
  \centering
  \caption{Main results on ImageNet-1k. \textbf{$*$} indicates the teacher is pre-trained on ImageNet-21K. We evaluate DeiT-B with the best performance because of the undulation when training the model.}
  \begin{tabular}{c|l|c|lc}
    \toprule
    Teacher& Student & Type  & Top-1 Accuracy & Top-5 Accuracy\\
    \midrule
    \multirow{5}{*}{\makecell{DeiT-Small\\(80.69)}}
    &DeiT-Tiny    &- & 74.42 &92.29\\
    &KD          &logit & 75.01 ({\color{red}+0.59}) &92.52\\
    &NKD         &logit & 75.48 ({\color{red}+1.06}) &92.72\\
    &\cellcolor{lightgray!45}Ours &\cellcolor{lightgray!45}feature &\cellcolor{lightgray!45}75.40 ({\color{red}+0.98}) &\cellcolor{lightgray!45}92.66\\
    &\cellcolor{lightgray!45}Ours+NKD    &\cellcolor{lightgray!45}feature+logit & \cellcolor{lightgray!45}\textbf{76.18} ({\color{red}+1.76}) &\cellcolor{lightgray!45}\textbf{93.14}\\
    \midrule
    \multirow{5}{*}{\makecell{DeiT III-Small$^{*}$\\(82.76)}}
    &DeiT-Tiny &- & 74.42 &92.29\\
    &KD        &logit & 76.01 ({\color{red}+1.59}) &93.26\\
    &NKD       &logit & 76.68 ({\color{red}+2.26}) &93.51\\
    &\cellcolor{lightgray!45}Ours &\cellcolor{lightgray!45}feature &\cellcolor{lightgray!45}76.06 ({\color{red}+1.64}) &\cellcolor{lightgray!45}93.16\\
    &\cellcolor{lightgray!45}Ours+NKD  &\cellcolor{lightgray!45}feature+logit &\cellcolor{lightgray!45}\textbf{77.78} ({\color{red}+3.36}) &\cellcolor{lightgray!45}\textbf{93.97}\\
    \midrule
    \multirow{5}{*}{\makecell{DeiT III-Base$^{*}$\\(85.48)}}
    &DeiT-Small &- & 80.55 &95.12\\
    &KD        &logit & 82.52 ({\color{red}+1.97}) &96.30\\
    &NKD       &logit & 82.74 ({\color{red}+2.19}) &96.33\\
    &\cellcolor{lightgray!45}Ours &\cellcolor{lightgray!45}feature &\cellcolor{lightgray!45}81.95 ({\color{red}+1.40}) &\cellcolor{lightgray!45}95.64\\
    &\cellcolor{lightgray!45}Ours+NKD  &\cellcolor{lightgray!45}feature+logit &\cellcolor{lightgray!45}\textbf{83.59} ({\color{red}+3.04}) &\cellcolor{lightgray!45}\textbf{96.69}\\
    \midrule
    \multirow{5}{*}{\makecell{DeiT III-Large$^{*}$\\(86.81)}}
    &DeiT-Base &- & 81.76 &95.81\\
    &KD        &logit & 84.06 ({\color{red}+2.30})&96.77\\
    &NKD       &logit & 84.96 ({\color{red}+3.20}) &97.17\\
    &\cellcolor{lightgray!45}Ours &\cellcolor{lightgray!45}feature &\cellcolor{lightgray!45}83.46 ({\color{red}+1.70}) &\cellcolor{lightgray!45}96.41\\
    &\cellcolor{lightgray!45}Ours+NKD  &\cellcolor{lightgray!45}feature+logit &\cellcolor{lightgray!45}\textbf{85.41} ({\color{red}+3.65}) &\cellcolor{lightgray!45}\textbf{97.39}\\
    \bottomrule
  \end{tabular}
  \vspace{2pt} 
  \label{table:main}
\end{table*}

\subsection{Downstream task}
The model with ViTKD achieves significant improvements for the classification task on ImageNet. To further evaluate the effectiveness of the model with distillation, we try to apply the model to object detection. We use Mask-RCNN~\citep{he2017mask} as the detector and follow the training setting from ViTDet~\citep{li2022exploring} on detectron2~\citep{wu2019detectron2}.

As the results shown in Table~\ref{table:detection}, the backbone DeiT-Small trained with ViTKD brings the detector 1.21 mAP gains. When the backbone is trained with ViTKD and NKD, the mAP improvement can be boosted to 1.62. The results demonstrate the model trained with distillation has not only better performance for image classification but also stronger semantic information for the downstream task.

\begin{table*}[t]
\setlength{\tabcolsep}{7 pt}
  \centering
  \caption{The detection results on COCO. We use Mask-RCNN as the detector.}
  \begin{tabular}{c|ll|ll}
    \toprule
    \multirow{2}{*}{Distillation}&
    \multicolumn{2}{c|}{DeiT-Small}&
    \multicolumn{2}{c}{DeiT-Base} \\
    \cmidrule{2-5}
    &AP$^{box}$ &AP$^{mask}$ &AP$^{box}$ &AP$^{mask}$\\
    \midrule
    - &45.07&40.14&47.23 &41.88\\
    ViTKD &46.28 ({\color{red}+1.21})&41.05 ({\color{red}+0.91})&48.13 ({\color{red}+0.90})&42.82 ({\color{red}+0.94})\\
    ViTKD+NKD &46.69 ({\color{red}+1.62})&41.38 ({\color{red}+1.24})&48.83 ({\color{red}+1.60})&43.19 ({\color{red}+1.31})\\
    \bottomrule
  \end{tabular}
  \vspace{2pt} 
  \label{table:detection}
\end{table*}

\section{More Analyses}
\label{sec:analyses}

\subsection{Do we need to distill the middle layers?}
We have discussed the distillation strategies for the shallow and deep layers of ViT-based models. As shown in Figure~\ref{fig:attn_tiny} and~\ref{fig:attn_compare}, the attention distributions of the middle layers are similar to that of shallow layers. Accordingly, we explore the effects of distillation on the middle layers (\eg, the $6_{th}$ layer) in Table~\ref{table:middle layer}. In general, distillation on either the shallow or middle layers can benefit the student. The first layer's knowledge boosts the student most. Besides comparing the improvements from different distillation layers, we find that the knowledge from the shallow layers is much more helpful than that from the middle layer for distillation. Furthermore, when combing the shallow and middle layers together, the accuracy improvement is just $0.02\%$. Considering the trade-offs between time consumption and performance, we do not distill on the middle layers eventually.

\begin{table}[h]
\centering
\begin{minipage}[t]{0.48\textwidth}
	\renewcommand\arraystretch{1.3}
	\setlength\tabcolsep{0.7mm}
	\centering
  \caption{The effect of distillation on different layers by a \emph{Mimicking} way.}
  \begin{tabular}{c|cccccc}
  \toprule
    Layer & \multicolumn{6}{c}{DeiT Small -- DeiT Tiny}\\
    \midrule
    0 & - &\checkmark&-&-&\checkmark&\checkmark\\
    1 & - &-&\checkmark&-&\checkmark&\checkmark\\
    6 & - &-&-&\checkmark&-&\checkmark\\
    \midrule
    Top-1 & 74.42&75.01&74.97 &74.72&75.12&75.14\\
    \bottomrule
  \end{tabular}
  \label{table:middle layer}
  \end{minipage}
  \hspace{2mm}
\begin{minipage}[t]{0.48\textwidth}
	\renewcommand\arraystretch{1.3}
	\setlength\tabcolsep{2.0mm}
	\centering
  \caption{The Top-1 accuracy comparisons of using different teachers to distill the student DeiT-Tiny's shallow layers via mimicking.}
  \begin{tabular}{c|c|l}
    \toprule
    Method& Teacher & \quad Student\\
    \midrule
    baseline & -&74.42\\
    CaiT-S24&83.37  & 73.12 ({\color{green}-1.30})\\
    DeiT-Small&80.69  & 75.12 ({\color{red}+0.70})\\
    DeiT III-Small&82.76  & 75.31 ({\color{red}+0.89})\\
    \bottomrule
  \end{tabular}
  \label{table:different layers}
  \end{minipage}
\end{table} 

\subsection{Teachers with the same architecture as the student are appropriate.}
\label{sec:better t}
\begin{figure}[t]
    \centering
    \includegraphics[width=\textwidth]{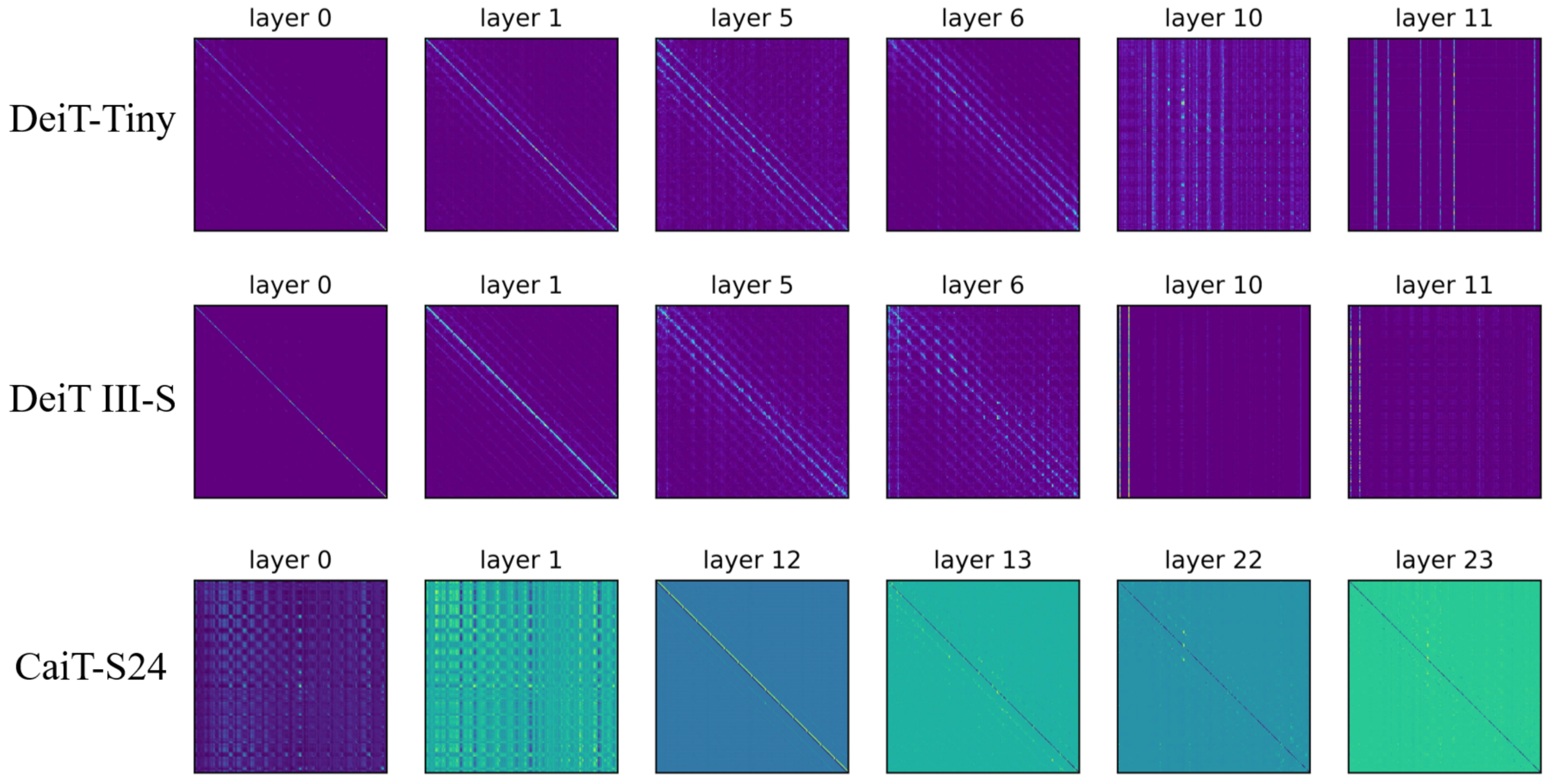}
    \caption{Visualization of the average attention map from the student (DeiT-Tiny) and two different teachers (DeiT III-Small and CaiT-S24).}
    \label{fig:attn_compare}
\end{figure}

We have chosen teachers with the same architecture to guide the student in Table~\ref{table:main}, achieving significant improvements. 
In this subsection, we explore whether a teacher with different architecture is still suitable for distillation. Here we choose CaiT-S24~\citep{Touvron_2021_ICCV} as the teacher, which has a high performance of $83.37\%$ and a different architecture with DeiT. 
Following the previous guidance, we utilize CaiT-S24's shallow layers to distill the student's shallow layers, which leads to a $1.29\%$ accuracy drop (shown in Table~\ref{table:different layers}). To further analyze what causes the degradation, we visualize the attention maps of the three used models in Figure~\ref{fig:attn_compare}. Interestingly, the shallow and deep layer's attention distributions of DeiT and CaiT are quite different, making it hard for the student to learn such attention. This phenomenon is not consistent with the assumptions of our proposed ViTKD, which causes inevitable performance degradation. In contrast, when using a teacher with the same architecture for distillation, the student gains noticeably. This observation indicates that a teacher with the same architecture who generates similar attention, is more suitable for ViTKD.

\subsection{Effects of different losses}

As described in the practical guidance, we distill the shallow layers and deep layers by mimicking and generation, respectively. In this subsection, we conduct experiments of \emph{Mimicking} loss $\mathcal{L}_{lr}$ and \emph{Generation} loss $\mathcal{L}_{gen}$ to investigate their influences on the student with DeiT-Tiny. As shown in Table~\ref{table:ablation study}, both the knowledge from shallow and deep layers are helpful for the student. When just applying a single loss, the $\mathcal{L}_{lr}$ on shallow layers benefits the student much more than $\mathcal{L}_{gen}$ on the deep layer. This phenomenon shows that incipient attention knowledge really matters for ViT's feature distillation, which is completely different from the CNN-based model's feature distillation. Furthermore, these two losses are complementary to each other. For example, when combing $\mathcal{L}_{lr}$ and $\mathcal{L}_{gen}$ together, the student with a DeiT III-Small teacher achieve $76.06\%$ Top-1 Accuracy, which is much higher than just applying $\mathcal{L}_{lr}$'s $75.31\%$ and $\mathcal{L}_{gen}$'s $74.79\%$.

\begin{table}[t]
\setlength\tabcolsep{15 pt}
  \centering
  \caption{Ablation study of the losses of \emph{Mimicking} and \emph{Generation} distillation.}
  \begin{tabular}{c|c|ccc}
    \toprule
    Losses & \multicolumn{4}{c}{DeiT-Tiny (Student)}\\
    \midrule
    $\mathcal{L}_{lr}$  & - &\checkmark&-&\checkmark\\
    $\mathcal{L}_{gen}$ & - &-&\checkmark&\checkmark\\
    \midrule
    DeiT-Small (Teacher) & 74.42 &75.12&74.72&{\bf75.40}\\
    DeiT III-Small (Teacher) & 74.42 &75.31&74.79&{\bf76.06}\\
    \bottomrule
  \end{tabular}
  \label{table:ablation study}
\end{table}

\subsection{Sensitivity study of hyper-parameters}
\begin{figure}[t]
  \centering
  \begin{subfigure}{0.495\textwidth}
    \includegraphics[width=1\textwidth]{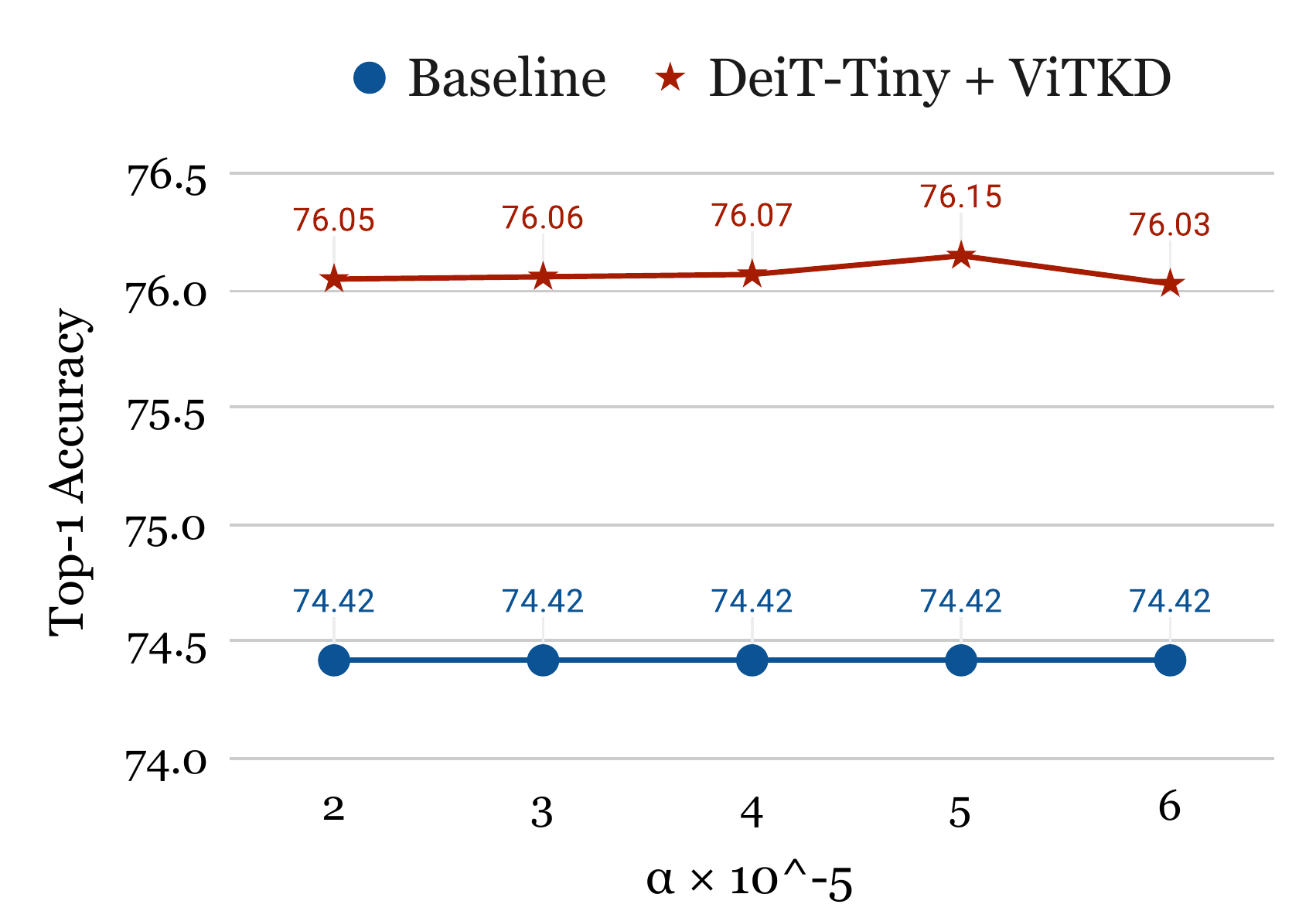}
    \caption{$\alpha$ for the shallow layers' loss $\mathcal{L}_{lr}$}
    \label{hyper-alpha}
  \end{subfigure}
  \begin{subfigure}{0.495\textwidth}
    \includegraphics[width=1\textwidth]{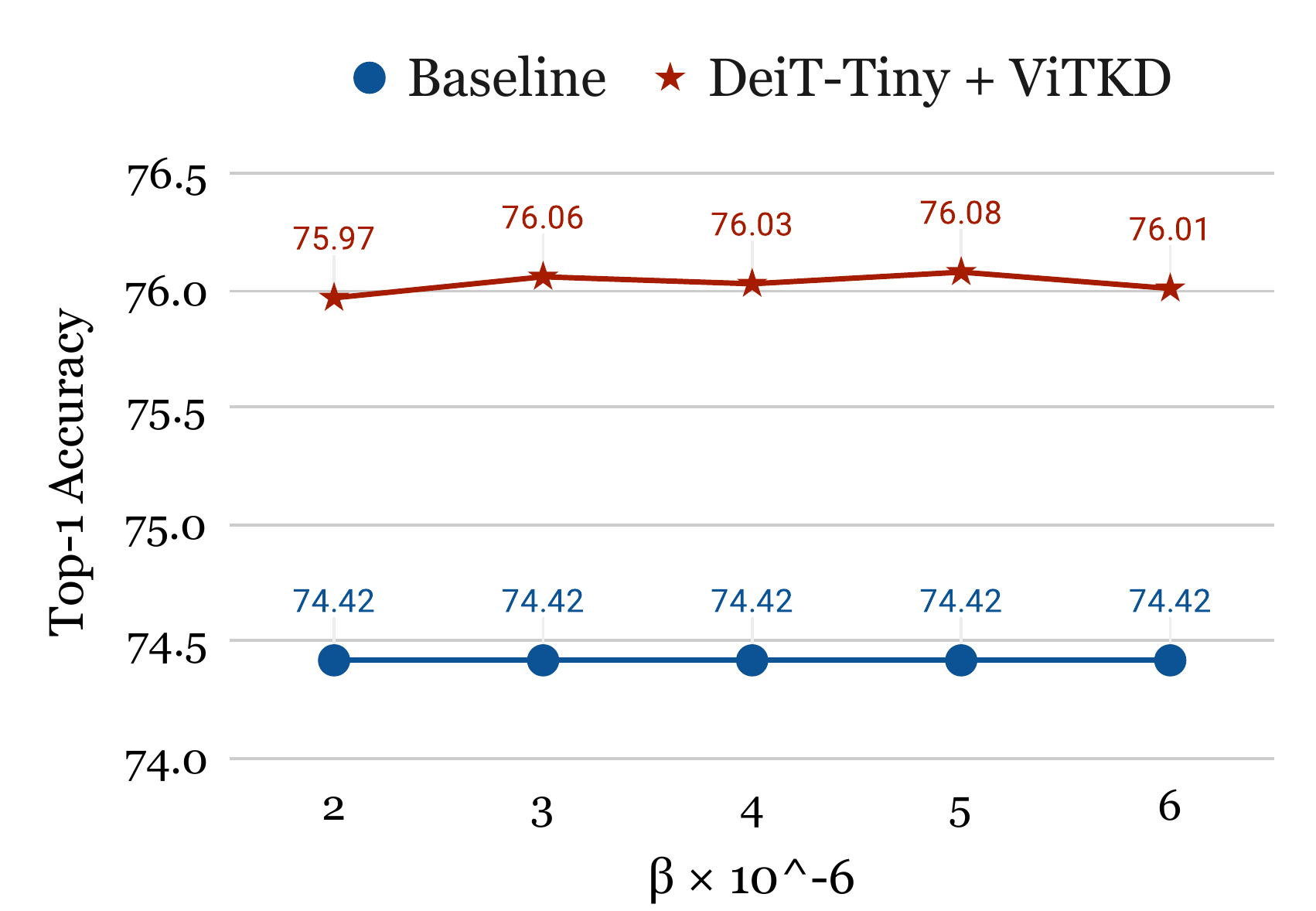}
    \caption{$\beta$ for the deep layers' loss $\mathcal{L}_{gen}$}
    \label{hyper-beta}
  \end{subfigure}
  \caption{The sensitivity study of hyper-parameters $\alpha$ (a) and $\beta$ (b).} 
  \label{fig:hyper}
\end{figure}

In ViTKD, we use $\alpha$ and $\beta$ in Equation~\ref{eq:all loss} to balance the shallow layer's distillation loss $\mathcal{L}_{lr}$ and the deep layer's distillation loss $\mathcal{L}_{gen}$, respectively. To explore the sensitivity of the hyper-parameters, we conduct experiments by adopting DeiT III-Small to distill DeiT-Tiny on ImageNet-1K.
As shown in Figure~\ref{fig:hyper}, ViTKD is not sensitive to the hyper-parameters $\alpha$ or $\beta$ which is just used for balancing the distillation loss. Specifically, when $\alpha$ varies from 2 to 6, the student's worst accuracy is $76.03\%$, which is just $0.12\%$ lower than the highest accuracy. Besides, it is still $1.61\%$ higher than the baseline model, demonstrating our ViTKD is not sensitive to the hyper-parameters.

\section{Conclusion}
\label{sec:con}

In this paper, we explore a feature-based distillation method for ViT-based models. To this end, we design a series of experiments and discuss the effects of different distillation methods, layers, and modules. From the results, We derive three practical guidelines for ViT's feature distillation. We propose our method ViTKD based on the guidelines, which includes the distillation on shallow layers via \emph{mimicking } and deep layers via \emph{generation}. ViTKD brings the student significant improvements on the image classification task and also benefits other downstream task. Besides, ViTKD is truly a feature-based method that can be easily combined with logit-based distillation methods to further improve the student.

\subsubsection*{Limitations}
We use the \emph{mimicking} method for the shallow layer's distillation and the \emph{generation} method for the deep layer's distillation. However, the way to achieve \emph{mimicking} and \emph{generation} is still simple and needs further exploration. Moreover, it is still interesting to transfer the feature knowledge to the student from a teacher with different architecture.


\newpage

\bibliography{iclr2023_conference}
\bibliographystyle{iclr2023_conference}


\end{document}